# Quantum Interference for Counting Clusters

Rohit R Muthyala   Davi Geiger   Zvi M. Kedem

**Abstract**

Counting the number of clusters, when these clusters overlap significantly is a challenging problem in machine learning. We argue that a purely mathematical quantum theory, formulated using the path integral technique, when applied to non-physics modeling leads to non-physics quantum theories that are statistical in nature. We show that a quantum theory can be a more robust statistical theory to separate data to count overlapping clusters. The theory is also confirmed from data simulations.

This works identify how quantum theory can be effective in counting clusters and hope to inspire the field to further apply such techniques.

## 1  Introduction

The problem of finding the number of clusters in a given data set is of importance in machine learning. Most techniques of clustering assume the number of clusters to be known. Techniques to estimate the number of clusters often bundle the problem with all the other cluster parameters estimation, such as finding the number of points associated to each cluster, their centers, and other cluster parameters that best fit the given data, according to some model described by some optimization criteria. Bundling the problems makes more difficult to resolve. Moreover, very little examination exists to the resistance of the methods to detect multiple clusters with significant spatial overlapping. For example, if we do have a distribution of the data to come from two Gaussian distributions, with centers $\mu_1$ and $\mu_2$ and both with covariance $\Sigma$, we can describe the data by the distribution

$$\mathrm{P}_0(x) = n_1\, G_\Sigma(x - \mu_1) + (1 - n_1)\, G_\Sigma(x - \mu_2) \qquad (1)$$

where $n_1$ is the ratio between the number of points drawn from the first Gaussian distribution, $G_\Sigma(y - \mu_1)$, and the total number of points. We develop techniques to effectively detect multiple instances of a Gaussian model when the unit-less distance $\delta\mu = \frac{|\mu_2 - \mu_1|}{2\sqrt{\det \Sigma}}$ between any two Gaussian instance becomes smaller than 1. To illustrate in one dimension the problem, we plot (1) as the distance between the centers $\delta\mu = \frac{|\mu_2 - \mu_1|}{2\sigma}$ varies, where $\sigma^2$ is the variance (see figure 1). It is noticeable that for small values of $\delta\mu$ the two peaks merge into one peak. We investigate the use of a classical statistical method and a quantum method to count the two peaks in such overlapping scenarios.

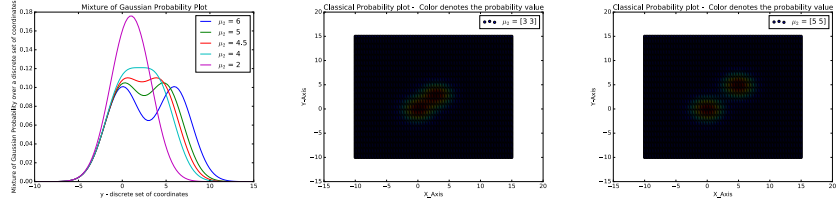

Figure 1: Data distributions of two clusters in 1D and 2D for P($x$) from (1), with $n_1 = 0.5$, $\sigma = 2$ and for 2D we have the same eigenvalues, and $\mu_1 = 0$ (or $(0,0)$ in 2D). For (b) we vary $\mu_2 = 2, 4, 4.5, 5, 6$ (or $\delta\mu = \frac{|\mu_2 - \mu_1|}{2\sigma} = 0.5, 1, 1.125, 1.25, 1.5$). For (b) we show $\mu_2 = (3, 3)$ and (c) $\mu_2 = (5, 5)$. The two peaks are readily visible for $\mu_2 = 4.5, 5, 6$, but they merge at $\mu_2 = 2, 4$ and so demonstrate the challenge of counting the clusters when they significantly overlap. The problem described here in 1D and 2D for two clusters is representative of a multi-clustering problem at higher dimensions where data from overlapping multi-clustering pose a challenge to accurately recover the number of clusters.

Our model start from the classical optimization criteria associated to a Gaussian data generator and is given by half the square of the Mahalonobis distance

$$\mathcal{A}(x, y, T) = \frac{1}{2} (x - y)^\mathsf{T} \eta^{-1}(T) (x - y) . \quad (2)$$

where $y$ is the center to data described by $x$ and $T > 0$ is a parametrization (or scaling) of the covariance $\eta(T)$, the larger is $T$ the larger are the eigenvalues. Each Gaussian in (1) is then a Gibbs distribution of an instance of the optimization model (2). We focus on scenarios where multiple data instances of the model are present, the number of them is unknown, and they overlap. We are seeking stable detection methods, that have consistency on detecting the number of minima as models instances overlap. The condition for stability is that near the minima the model's response decay fast and so multiple overlapping (and noisy) cluster data will all be detected.

We obtain stable detection by applying quantum techniques to the classical modeling and we compare it to a classical statistical model counterpart. Our approach is to use mathematical tools extracted from quantum mechanics to the counting cluster problem and to study the extent of the applicability. We stress that we extract a set of mathematical tools from quantum mechanics, but we do not use physics modeling (e.g., the Schrödinger equation is never considered [1].)

The key observation that led us to the quantum methods is the relevance of the quantum *interference phenomenon*, not present in classical methods. Due to interference, false probability hypotheses *can weaken or cancel each other* and thus the true hypothesis emerges with greater saliency. That is the motivation behind our approach, and here we develop the formalism to demonstrate its effectiveness.

---

[1] Schrödinger equation is obtained from the quantum path integral method when the physical energy model is given by the classical kinetic energy and a potential energy. We never consider such physical modeling.

## 1.1 Previous Work and the Originality of Our Approach

There has been extensive research conducted in ML dedicated to make clustering methods robust, such as the Gaussian Mixture model [1], which yields good parameter detection when the number of clusters is known and the overlap between them is not too large. Here we explore scenarios where the number of clusters is unknown and clusters overlap significantly with each other.

Proposal to count the number of clusters are integrated with the detection of all other cluster parameters. A state of the art approach is an information–theoretic approach based on the rate distortion theory called the "jump" method [6]. The strategy of the algorithm is to generate a distortion curve for the input data by running a standard clustering algorithm such as k-means for all values of k between 1 and n, and computing the distortion (described below) of the resulting clustering. Since the number of unknowns is large, and the algorithm attempts to solve for all the clustering unknowns simultaneously, such method will fail for clusters with larger overlaps. In contrast, we focus on the detection of the number of clusters, in the presence of overlaps, without accurate investigation of the other cluster parameters.

We propose a novel approach: development of algorithms based on *purely mathematical properties* underlying quantum mechanics.

Quantum-based algorithms were used outside physics (e.g., [5, 4, 8], both on clustering algorithm using quantum mechanics) and the Schrödinger equation from physics was employed. Therefore, roughly speaking, problems outside physics that had no connection to *quantum mechanics* were addressed using quantum-mechanics behavior. In contrast, we propose to use mathematical techniques to CV modeling and so the Schrödinger equation is not considered. We use the purely mathematical technique of *path integrals*, to model and solve optimization problems in ML. The trust of this proposal is that quantum-based and not physics-based methods (i) leads to more robust methods in ML and (ii) are general and so can be used in various problems and applications.

Quantum clustering algorithms were proposed, e.g. [4]. More recently quantum clustering algorithms were proposed by [8]. However, they are rooted on the Schrödinger equation, are quadratic in the number of points, and have difficulties with overlapping clusters. Another suggestion of using statistical quantum methods for clustering have been proposed [7]. However, this method is based on a mixture of quantum states, which is not a "pure quantum state". Statistical Quantum mixtures are not derived from the magnitude square of the wave function and do not exhibit interference, which is a central property of our formulation. So none of the phenomenon and formulas used here, are described by quantum mixture of states. That is not to say that Statistical Quantum mixtures cannot be useful, but they will not have the useful properties that we describe and exploit here. Not surprisingly, none of these methods have shown to perform on the task of counting clusters. In fact, maybe their main weakness is that they have not performed well on any task in clustering. They all lack a demonstration of what advantage there is in quantum theory to estimate any parameter associated to the clustering problem.

Our approach is quite different and we demonstrate its effectiveness. We start from the classical clustering method, which requires linear computations in the number of points. We derive the quantum probability amplitude as a pure state which is also computationally linear in the number of points. We demonstrate how quantum interference resolves the separation of data to count clusters in challenging scenarios of significant overlapping clusters.

## 2 Empirical Probability

We are given a set of input data points $X = \{x_i; i = 1, \ldots N\}$ and so the empirical continuous probability density is given by

$$P_0(x) = \frac{1}{N} \sum_{i=1}^{N} \delta(x - x_i),$$

where $\delta(x - x_i)$ is the Dirac distribution. The quantum method, and corresponding classical method, to evaluate the probability for the centers is derived in appendix A yielding

$$P_T^c(y) = \frac{1}{Z} \int dx \, e^{-\alpha \mathcal{A}(x,y,T)} P_0(x).$$

$$\psi_T(y) = \frac{1}{\sqrt{C}} \int dx \, e^{\frac{i}{\hbar} \mathcal{A}(x,y,T)} \psi_0(x),$$

where $\psi_0(x)$ is an initial probability amplitude derived from $P_0(x)$ and we assign an unknown phase term $e^{i\varphi_i}$ to each data point. The quantum probability is given by $P_T^q(y) = |\psi_T(y)|^2$. Note that the probability amplitude is a complex valued function. When the data probability $P_0(x)$ comes from multiple instances of the model, each instance may be assigned a phase, that will need to be estimated. We refer to $K^c(x, y, T) = \frac{1}{Z} e^{-\alpha \mathcal{A}(x,y,T)}$ as the classical statistical kernel and $K^q(x, y, T) = \frac{1}{\sqrt{C}} e^{\frac{i}{\hbar} \mathcal{A}(x,y,T)}$ as the quantum kernel. Typically, the classical kernel is a smoothing filter where the parameter $\alpha$ controls the amount of smoothness. Inserting the empirical data probability 2 to the classical and quantum model we obtain

$$P_T^c(y) = \frac{1}{Z} \int dx \, e^{-\alpha \mathcal{A}(x,y,T)} \frac{1}{N} \sum_{i=1}^{N} \delta(x - x_i) = \frac{1}{NZ} \sum_{i=1}^{N} e^{-\alpha \mathcal{A}(x_i,y,T)}$$

$$\psi_T(y) = \frac{1}{\sqrt{NC}} \sum_{i=1}^{N} e^{\frac{i}{\hbar} \mathcal{A}(x_i,y,T)} e^{i\varphi_i}$$

$$P_T^q(y) = |\psi_T(y)|^2 = \frac{1}{NC} \sum_{i=1}^{N} \sum_{j=1}^{N} e^{\frac{i}{\hbar}[\mathcal{A}(x_i,y,T) - \mathcal{A}(x_j,y,T)]} e^{i(\varphi_i - \varphi_j)}$$

$$= \frac{1}{NC} \sum_{i=1}^{N} \left( 1 + 2 \sum_{j>i}^{N} \cos\left( \frac{1}{\hbar} [\mathcal{A}(x_i, y, T) - \mathcal{A}(x_j, y, T)] + [\varphi_i - \varphi_j] \right) \right) \quad (3)$$

where typically, $\hbar$ is of the order of the action difference $\left[\mathcal{A}(x_i, y, T) - \mathcal{A}(x_j, y, T)\right]$ for pairs $(x_i, x_j)$ that interfere positively, thus, increasing the probability. For pairs such that their action difference is larger, so larger phase, the interference is negative reducing the final probability. The phase contribution $\varphi_i - \varphi_j$ adjusts the regime of negative and positive interference. This interference phenomenon is not present in classical systems.

## 2.1 A Gaussian Mixture Model

Data coming from Gaussian distributions are associated to the action given by (2). We do not know a priori the number of clusters present in a data set. Then, applying this model (2) to the classical probability and the quantum probability (3) we obtain

$$P_T^c(y) = \frac{1}{D} \sum_{i=1}^{N} G_{\frac{1}{\alpha}\eta}(x_i - y) \tag{4}$$

$$\psi_T(y) = \frac{1}{C} \sum_{i=1}^{N} G_{i\hbar\eta}(x_i - y)\, e^{i\varphi_i} \tag{5}$$

$$P_T^q(y) = \frac{1}{d} \sum_{i=1}^{N} \left[ \left( 1 + 2 \sum_{j>i}^{N} \cos\left(\phi_{ij}(y) + (\varphi_i - \varphi_j)\right) \right) \right] \tag{6}$$

where $D, C, d$ are the normalization constants,

$$\phi_{ij}(y) = -(x_i - x_j)^\top (\hbar\eta)^{-1} y + \frac{1}{2}\left(x_i^\top (\hbar\eta)^{-1} x_i - x_j^\top (\hbar\eta)^{-1} x_j\right),$$

and the hyper-parameters $\alpha$ and $\hbar$ scale the covariance $\eta$. For the one dimensional case the covariance $\eta$ becomes the variance, say $\lambda^2$, and the estimation of the hyper-parameter is bundled with the estimation of the variances $\lambda^2$.

### 2.1.1 Experiments with Two Gaussian Instances

We simulate multiple data in 1D generated by two Gaussian distributions, $G_\Sigma(x - \mu_1)$ and $G_\Sigma(x - \mu_2)$, one centered at $\mu_1 = 0$ and the other varying the position of $\mu_2$ and so $\delta\mu = \frac{\mu_2}{2\sqrt{\det \Sigma}}$. We chose two Gaussian distributions with the same covariance and same number of points drawn, but different centers. The choice is to help us focus on the problem of detecting the number of clusters, i.e., number of centers, and so we do not want any information from the other parameters to facilitate such a task. For example, if the covariances are different one may detect two clusters by tunning to detect such a difference. Thus, from the each Gaussian distribution we sample a set of $\frac{N}{2}$ data points from each Gaussian, producing the data $X = \{x_i; i = 1, \ldots N\}$. We then compute the probabilities (5) and (6) and investigate the detection of the two centers as the overlap $\delta\mu$ vary.

In order to ensure that the overlap is large, we examine the parameters of the two Gaussian models where the overlap cause the two peaks to merge, shown in figure 1. We choose centers $\mu_1 = 0$ and $\mu_2 = 4$, both with standard deviation $\sigma = 2$. For the classical statistical method, we examine the values of $\alpha$ that may help detect the two peaks associated to the two clusters. Note that the statistical classical method describes the data smoothed by a Gaussian filter of variance $\frac{1}{\alpha}\sigma^2$ (see fgure 2). The larger the variance the more smoothing, and so the two peaks are less likely to be detected. In contrast, when $\alpha \to \infty$ we end up with data points that coincide with the discrete set of $y$ values and all of them will be peaks. We show empirically that as we vary $\alpha$ there is an impossibility to detect two peaks.

In contrast, the quantum probability is capable to detect the two peaks, quite easily with phase value that cause negative quantum interference between different cluster data (see figure 3 Left). In fact, even with a random assignment of phases to each data point, the two peaks are revealed (see figure 3 Right).

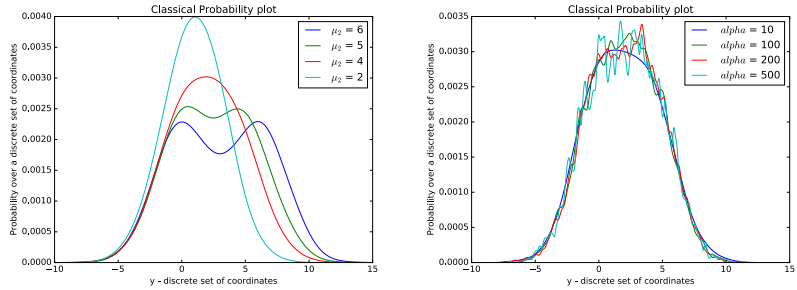

Figure 2: Data is drawn from a 1D mixture of two Gaussian distributions, each with 3000 points, centered at $\mu_1 = 0$ and $\mu_2 = 6, 5, 4, 2$, both with standard deviation $\sigma_1 = \sigma_2 = 2$. Left: Classical probability plot discretized over 100 intervals over y with $\alpha = 10$. Right: For $\mu_2 = 4$ and various large values of $\alpha$, less smoothing, and one cannot find two peaks corresponding to two centers.

## 3   Analysis of the Probabilities

We now derive theoretical results supporting the previous section's experiments. Our focus is to compare the classical and quantum probabilities, in particular how fast the probability decay from its maximum, which in turn predicts the ability to count the number of clusters with overlaps. For Gaussian distributions the decay from the maximum is measured by the variances or equivalently relative entropy. The smaller is the variance, the lower is the entropy, the faster is the decay from the maximum.

We investigate the stability and ability to detect the centers in case where the data is drawn from two overlapping Gaussian distributions. It establish the advantage of using quantum theory to detect two clusters when they overlap significantly. The lower entropy of the quantum system provides more stability of the solution near the cluster centers $\mu_1$ and $\mu_2$. The detection of the two centers is more likely to occur via the quantum probability.

**Lemma 1** (Two Model Instances in 1D). *Given an action $\mathcal{A}(x, y, T)$ described by* (2),

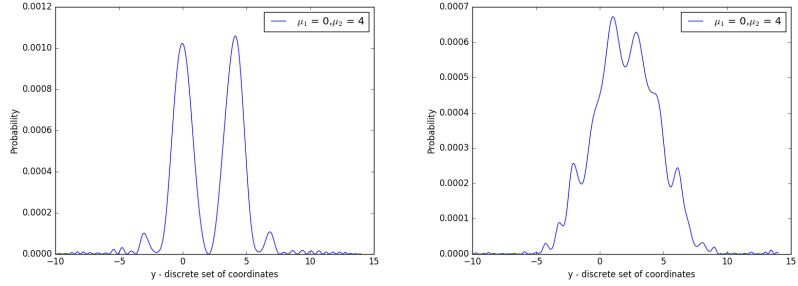

Figure 3: 1D Quantum probabilities vs locations for data generated used in figure 2 for $\hbar = 0.4$ (which is just larger than $\frac{1}{\pi}$, as required from lemma 1). Left: we assign phase $\phi_1 = 0$ to the data drawn from the first Gaussian centered at $\mu_1 = 0$ and $\phi_2 = \pi$ for the Gaussian centered at $\mu_2 = 0$. Right: We assign a random phase to each data point, and so no knowledge of the data is used. For both scenarios, a simple threshold of the probabilities to 1/2 of the maxima probability followed by the detection of the maxima, we obtain the correct counting of two peaks. It is beyond the scope of this work to investigate the use of this method to detect cluster center locations.

*where $x, y$ are coordinates in 1 dimension. Assume the data is drawn from two model instances overlapping, with centers $\mu_1$ and $\mu_2 = \mu_1 + \Delta\mu$ and both with variance $\sigma^2$. Each model instance yields a unique solution $y_1$ and $y_2$, i.e., $y_1 = \arg\min_y \mathcal{A}(\mu_1, y, T)$ and $y_2 = \arg\min_y \mathcal{A}(\mu_2, y, T)$ are both local minimum.*

*Then, near the minima and for $\hbar > \frac{1}{\pi}$, the lowest entropy of the quantum probability, $\mathcal{S}(\mathrm{P}_T^q(y); \hbar)$ (there is a choice of phases to be made), is lower than the entropy of the classical probability $\mathcal{S}(\mathrm{P}_T^c(y); \alpha)$ for any choice of $\alpha$. Moreover, the detection of two model instances, with overlap $\delta\mu = \frac{|\Delta\mu|}{2\sigma} \approx 1$, is obtained straightforwardly from $\mathrm{P}_T^q(y)$ via thresholding and counting its maxima, while such a simple method does not work for the classical probability.*

*Proof.* Each data distribution is a Gaussian described by $G_\Sigma(x - \mu_1)$ and $G_\Sigma(x - (\mu_1 + \Delta\mu))$. Each one is associated to a quantum probability amplitude, $\sqrt{G_\sigma(x - \mu_1)}\,\mathrm{e}^{\mathrm{i}\phi_1}$ and $\sqrt{G_\sigma(x - \mu_2)}\,\mathrm{e}^{\mathrm{i}\phi_2}$. Let us consider $P$ points are drawn altogether from these two distributions, where $0 < n_1 < 1$ is the proportion of points drawn from the Gaussian $G_\Sigma(x - \mu_1)$. Thus, $1 - n_1$ is the proportion of points drawn from the Gaussian distribution $G_\sigma(x - \mu_2)$. We combine them to write the total quantum probability amplitude $\psi_0(x) = \frac{1}{c}\left[\sqrt{n_1}\sqrt{G_\sigma(x - \mu_1)}\,\mathrm{e}^{\mathrm{i}\phi_1} + \sqrt{(1 - n_1)}\sqrt{G_\sigma(x - (\mu_1 + \Delta\mu))}\,\mathrm{e}^{\mathrm{i}\phi_2}\right]$, where $c$ is a normalization. Similarly, for the classical statistical case we have $\mathrm{P}_0(x) = n_1 G_\sigma(x - \mu) + (1 - n_1)G_\sigma(x - (\mu + \Delta\mu))$, as described in (1).

The kernel for the classical and quantum cases are also Gaussian and so the classical probability (9) and the probability amplitude (10) become

$$P_T^c(y) = \int dx\, G_{\frac{1}{\sqrt{\alpha}}\lambda}(y-x)\left[n_1 G_\sigma(x-\mu_1) + (1-n_1)G_\sigma(x-(\mu_1+\Delta\mu))\right],$$

$$= [n_1 G_{\sigma_\alpha}(y-\mu_1) + (1-n_1)G_{\sigma_\alpha}(y-(\mu_1+\Delta\mu))],$$

$$\psi_T(y) = \int dx\, G_{\sqrt{i\hbar}\lambda}(y-x)\frac{1}{c}\left[\sqrt{n_1}\sqrt{G_\sigma(x-\mu_1)}\,e^{i\phi_1} + \sqrt{(1-n_1)}\sqrt{G_\sigma(x-(\mu_1+\Delta\mu))}\,e^{i\phi_2}\right]$$

$$= \frac{1}{c}\left[\sqrt{n_1}G_{\sigma_{i\hbar}}(y-\mu_1)\,e^{i\phi_1} + \sqrt{(1-n_1)}G_{\sigma_{i\hbar}}(y-(\mu_1+\Delta\mu))\,e^{i\phi_2}\right]$$

$$P_T^q(y) = |\psi_T(y)|^2$$

$$= \frac{n_1}{c^2}G_{\sigma_\hbar^\mathbb{R}}(y-\mu_1) + \frac{(1-n_1)}{c^2}G_{\sigma_\hbar^\mathbb{R}}(y-(\mu_1+\Delta\mu))$$

$$+ 2\frac{\sqrt{n_1(1-n_1)}}{c^2} G_{\sigma_\hbar^\mathbb{R}}\left(y-\left(\mu_1+\frac{\Delta\mu}{2}\right)\right) G_{\sqrt{2}\sigma}(\Delta\mu)\cos\left(-\frac{(\Delta\mu)^2}{2\hbar\lambda^2}+(\phi_1-\phi_2)\right)$$

where $\sigma_\alpha^2 = \sigma^2 + \frac{1}{\alpha}\lambda^2$, $\sigma_{i\hbar}^2 = \sigma^2 + i\hbar\lambda^2$, $(\sigma_\hbar^\mathbb{R})^2 = \frac{\sigma^2}{2} + \frac{\hbar^2\lambda^4}{2\sigma^2}$. Above we did various manipulations with product of two Gaussian, e.g., see [2] for derivations of such general manipulations.

The first two terms of the quantum probability are described by Gaussian with variance $(\sigma_\hbar^\mathbb{R})^2 = \frac{\sigma^2}{2} + \frac{\hbar^2\lambda^4}{2\sigma^2}$, while the two classical probability Gaussian have variance $\sigma_\alpha^2 = \frac{1}{\alpha}\lambda^2 + \sigma^2$. Since the relative entropy of a Gaussian distribution with variance $\sigma$ is $K + \frac{1}{2}\ln\sigma$, where $K$ is a constant, classical and quantum Gaussian terms have larger entropy than their associated data term. The classical model increases entropy as it smooth the data. The first two quantum terms do just the same. In fact, by equating $\frac{\hbar^2\lambda^4}{2\sigma^2} = \frac{\sigma^2}{2} + \frac{1}{\alpha}\lambda^2$ guarantees the first two quantum terms to have the same Gaussian behavior than the classical distribution. This smoothing process may be justified to reduce data sampling limitations. The more data is collected, the lesser need for smoothing and the smaller should $\frac{1}{\alpha}\lambda^2$ be. However, the quantum model excels due to the interference term, the third term. We can choose $\hbar$ so that $\pi \geq \left|-\frac{(\Delta\mu)^2}{2\hbar\lambda^2}+\phi_1-\phi_2\right| > \pi/2$, and the cosine becomes negative and so the third term Gaussian contributes negatively with the same variance as the first two terms and centered in between the two centers. In this case we can set $\phi_1 - \phi_2 = \pi$ and then $\frac{(\Delta\mu)^2}{2\hbar\lambda^2} < \frac{\pi}{2} \to \hbar > \frac{(\Delta\mu)^2}{\pi\lambda^2}$. For $\Delta\mu \approx \lambda$ we require $\hbar > \frac{1}{\pi}$. There is also a multiplication factor $G_{\sqrt{2}\sigma}(\Delta\mu)$, which is more significant for larger overlap scenarios ($\Delta\mu \approx \sigma$). When the overlap reduces, the multiplication factor decreases and the third term goes to zero, and the quantum method becomes equivalent to the classical method. Indeed in this scenario counting the number of clusters is straight forward by counting the number of maxima of the probability. The challenging scenario is when the overlapping is significant and when the quantum interference term recreates the two maxima of the original two Gaussian data distributions by subtracting a Gaussian term centered in between the two centers. This procedure also reduces the entropy and increases sparsity as measured by $\sqrt{P(y)}$.

We can then choose $\hbar$ so that the third term on the quantum probability, the interference term, to be negative, i.e., $\pi \geq \left|-\frac{(\Delta\mu)^2}{2\hbar\lambda^2} + \phi_1 - \phi_2\right| > \pi/2$ and $\frac{(\Delta\mu)^2}{2\hbar\lambda^2} < \frac{\pi}{2} \rightarrow \hbar > \frac{(\Delta\mu)^2}{2\pi\lambda^2}$. For $\Delta\mu \approx \lambda$ we require $\hbar > \frac{1}{2\pi}$. In order to reduce entropy, the first two terms of the quantum prefer as small as possible $\hbar$ to reduce smoothing. The third term prefers a larger $\hbar$ value to have a larger negative Gaussian term to recreate the two peaks from the two sources and thus, reduce entropy. Due to the third term the quantum method can reduce the entropy compared to the original data entropy while the classical method best case scenario is to

□

## 3.1 Multiple Gaussian Distribution in Higher Dimension

The results shown for two Gaussian and in 1D, 3.1, are readily generalized to multiple Gaussian in higher dimensions as follows

$$P_T^c(y) = \sum_{i=1}^{K} n_i G_{\Sigma_{i,\alpha}}(y - \mu_i)$$

$$\psi_T(y) = \frac{1}{c} \sum_{i=1}^{K} \sqrt{n_i} G_{\Sigma_{i,\hbar}}(y - \mu_i) e^{i\phi_i}$$

$$P_T^q(y) = \frac{1}{c^2} \sum_{i=1}^{K} n_i G_{\Sigma_{i,\hbar}^{\mathbb{R}}}(y - \mu_i)$$
$$+ \sum_{i=1}^{K} \sum_{j \neq i}^{K} \frac{\sqrt{n_i n_j}}{c^2} G_{\Sigma_{i,j,\hbar}^{\mathbb{R}}}(y - \mu_{ij}) G_{\Sigma_{i,j}}\left(\frac{\mu_i - \mu_j}{2}\right) \cos\left[\frac{(\mu_i - \mu_j)\Sigma_{i,j}^{-1}(\mu_i - \mu_j)}{\hbar^2} + (\phi_i - \phi_j)\right]$$

where $1 = \sum_{i=1}^{K} n_i$, and $\Sigma_{i,\hbar} = \Sigma_i + i\hbar\eta \left(\Sigma_{i,j,\hbar}\right)^{-1} = \left(\Sigma_{i,\hbar}\right)^{-1} + \left(\Sigma_{j,\hbar}^*\right)^{-1} \left(\Sigma_{i,\hbar}^{\mathbb{R}}\right)^{-1} = \left(\Sigma_{i,i,\hbar}\right)^{-1} \mu_{ij} = \Sigma_{i,j,\hbar}\left((\Sigma_{i,\hbar})^{-1}\mu_i + (\Sigma_{j,\hbar}^*)^{-1}\mu_j\right) \Sigma_{i,j} = \Sigma_{i,\hbar} + \Sigma_{j,\hbar}^* = \Sigma_i + \Sigma_j$. Even though there are multiple Gaussian mixtures, the quantum interference is a pair-wise phenomenon and so the analysis for two Gaussian is quite general.

## 4 Conclusion

Our observation that we put forward in this paper is that quantum theory can be, in some scenarios, a more robust statistical theory.

We showed this by considering problems of counting number of clusters when they significantly overlap.

We showed a lemma in one dimension that gives insights on why the quantum theory, with the interference phenomenon, helps to separate cluster data and thus, count the number of overlapping clusters.

The method is quite general and not exclusive for the clustering problem.

## A  Quantum Estimation

We start by describing briefly the quantum path integral method.

The paths are described by a set of state coordinates $\mathcal{P}_{t_0}^{t_0+T} = \{x(t): t \in (t_0, t_0+T)\}$ where $x(t)$ are states and they represent coordinates of points in $\mathbb{R}^N$ space. We may be given an assignment of the initial and final state coordinates, such as $x(t_0) = x_0$ and $x(t_0 + T) = x_T$. We will also consider a distribution over the initial state coordinate $x_0$, so various initial states are considered for one given final state $x_T$. We then create a functional $\mathcal{A}: \mathcal{P}_{t_0}^{t_0+T} \to \mathbb{R}$ (or $\mathcal{A}(\mathcal{P}_{t_0}^{t_0+T}) \in \mathbb{R}$) and such map is invariant by time translation of each path $\mathcal{P}_{t_0}^{t_0+T}$, i.e., it is independent of the starting time $t_0$, and only dependent on the time interval $T$. Thus, we refer to the paths as $\mathcal{P}^T$ and the functional is then represented by $\mathcal{A}(\mathcal{P}^T)$. We may expand the initial. Let us now consider the optimization problem of a given functional $\mathcal{A}(\mathcal{P}_{t_0}^{t_0+T})$ over all paths $\mathcal{P}_{t_0}^{t_0+T}$ in the time interval $[t_0, t_0 + T]$. In quantum path integral theory [3], the probability amplitude is constructed by considering all possible paths (satisfying the boundary conditions) with weights proportional to $e^{\frac{i}{\hbar}\mathcal{A}(\mathcal{P}^T)}$. More precisely, the probability amplitude $\psi_T(x_T)$ is obtained by integrating $\psi_0(x_0)$ over all paths of time length $T$ with each path weight proportional to $e^{\frac{i}{\hbar}\mathcal{A}(\mathcal{P}^T)}$ as follows

$$\psi_T(x_T) = \int dx_0 \int \frac{e^{\frac{i}{\hbar}\mathcal{A}(\mathcal{P}^T)}}{\sqrt{C}} \psi_0(x_0) \, d\mathcal{P}^T, \tag{7}$$

where $C$ is a normalization constant so that $1 = \int |\psi_T(x_T)|^2 \, dx_T$. The integration over all paths as well as the normalization may require technicalities to make it more precise and finite and we will address the meaning of such integration for our purposes. The quantum probability distribution is obtained as $\mathrm{P}_T^q(x_T) = |\psi_T(x_T)|^2$ and at $t_0$ it is $\mathrm{P}_0(x_0) = |\psi_0(x_0)|^2$. The coordinates $x_T$ describe the states of the system. The hyper-parameter $\hbar$ is the reduced Planck's constant in physics but in ML needs to be estimated. The functional $\mathcal{A}(\mathcal{P}^T)$ expresses the modeling of the problem at hand. Our proposal is for the action to model ML without relying on the equation that was developed for modeling mechanics.

We now turn to the classical statistical approach. Again, consider the optimization criterion $\mathcal{A}(\mathcal{P}^T)$, now to create a likelihood probability $\mathrm{P}_T(x_T|x_0) = \int d\mathcal{P}^T \frac{1}{Z} e^{-\alpha \mathcal{A}(\mathcal{P}^T)}$, where all the paths satisfy the boundary conditions, $x_0 = x(t_0)$ and $x_T = x(t_0 + T)$, in a time interval $T$. $Z$ is the normalization constant so that $1 = \int \mathrm{P}_T(x_T|x_0) \, dx_T$. Since the initial coordinate is distributed according to $\mathrm{P}_0(x_0)$, a Bayesian approach combines these probabilities as follows

$$\begin{aligned}\mathrm{P}_T^c(x_T) &= \int dx_0 \, \mathrm{P}_T(x_T|x_0) \, \mathrm{P}_0(x_0) \\ &= \int dx_0 \int d\mathcal{P}^T \frac{1}{Z} e^{-\alpha \mathcal{A}(\mathcal{P}^T)} \mathrm{P}_0(x_0).\end{aligned} \tag{8}$$

We summarized by saying we constructed a classical statistical probability (equation 8) and a quantum probability (the magnitude square of equation 7), both derived from the optimization $\mathcal{A}(\mathcal{P}^T)$.

Thus, our model is of the form $\mathcal{A}(\mathcal{P}^T) = \mathcal{A}(x(t_0), x(t_0 + T), T) \equiv \mathcal{A}(x, y, T)$, where $\mathcal{A}(x, y, T)$ is a function of the initial coordinate, the final coordinate, and the time interval. Then our propagation model (8) becomes

$$P_T^c(y) = \frac{1}{Z} \int dx\, e^{-\alpha \mathcal{A}(x,y,T)}\, P_0(x)\,. \tag{9}$$

Similarly, (7) becomes

$$\psi_T(y) = \frac{1}{\sqrt{C}} \int dx\, e^{\frac{i}{\hbar} \mathcal{A}(x,y,T)} \psi_0(x)\,, \tag{10}$$